\documentclass{article} 
\usepackage{iclr2025_conference,times}

\usepackage{amsmath,amsfonts,bm}

\def\eqref#1{equation~\ref{#1}}

\def\1{\bm{1}}

\DeclareMathAlphabet{\mathsfit}{\encodingdefault}{\sfdefault}{m}{sl}
\SetMathAlphabet{\mathsfit}{bold}{\encodingdefault}{\sfdefault}{bx}{n}

\usepackage{hyperref}
\usepackage{url}

\usepackage{amsmath} 
\usepackage{multirow}
\usepackage{tikz}
\usepackage{todonotes}
\usepackage{booktabs}
\usetikzlibrary{arrows.meta}
\usepackage{amsmath}

\title{ConDiff: A Challenging Dataset for Neural \\ Solvers of Partial Differential Equations}

\author{%
  Vladislav Trifonov$^{~1~2}$~~Alexander Rudikov$^{~3~1}$~~Oleg Iliev$^{~4}$\\ \textbf{Yuri M. Laevsky}$^{~5}$~~\textbf{Ivan Oseledets}$^{~3~1}$~~\textbf{Ekaterina Muravleva}$^{~2~1}$ \\ \\
  $^{1}$Skolkovo Institute of Science and Technology, Moscow, Russia \\
  $^{2}$Sberbank of Russia, AI4S Center, Moscow, Russian Federation \\
  $^{3}$Artificial Intelligence Research Institute (AIRI), Moscow, Russia \\
  $^{4}$Fraunhofer Institute for Industrial Mathematics ITWM, Kaiserslautern, Germany \\
  $^{5}$Institute of Computational Mathematics and Mathematical Geophysics SB RAS, \\~~Novosibirsk, Russia \\
  \texttt{vladislav.trifonov@skoltech.ru}
}

\iclrfinalcopy 
\begin{document}

\maketitle

\begin{abstract}
We present ConDiff, a novel dataset for scientific machine learning. ConDiff focuses on the parametric diffusion equation with space dependent coefficients, a fundamental problem in many applications of partial differential equations (PDEs). The main novelty of the proposed dataset is that we consider discontinuous coefficients with high contrast. These coefficient functions are sampled from a selected set of distributions. This class of problems is not only of great academic interest, but is also the basis for describing various environmental and industrial problems. In this way, ConDiff shortens the gap with real-world problems while remaining fully synthetic and easy to use. ConDiff consists of a diverse set of diffusion equations with coefficients covering a wide range of contrast levels and heterogeneity with a measurable complexity metric for clearer comparison between different coefficient functions. We baseline ConDiff on standard deep learning models in the field of scientific machine learning. By providing a large number of problem instances, each with its own coefficient function and right-hand side, we hope to encourage the development of novel physics-based deep learning approaches, such as neural operators, ultimately driving progress towards more accurate and efficient solutions of complex PDE problems.
\end{abstract}

\section{Introduction}
\label{sec:introduction}

In recent years, machine learning techniques have emerged as a promising approach to solving PDEs, offering a new perspective in scientific computing. Machine learning algorithms, especially those based on neural networks, have demonstrated success in approximating complex functions and physical phenomena. Neural networks can provide more efficient and scalable methods compared to traditional numerical methods, which can be computationally expensive and limited by the dimensionality of the problem to be solved. Approaches using physical losses~\citep{karniadakis2021physics}, operator learning~\citep{li2020fourier}, symmetries incorporation~\citep{wang2020incorporating}, data-driven discretization~\citep{bar2019learning} lead to more physically meaningful solutions and gave neural networks better recognition than just black-boxes. 

Classical methods for solving PDEs have been extensively developed and refined over the years, providing a basis for understanding and analyzing various physical phenomena. These methods involve discretization the PDEs using techniques as the finite difference method~\citep{leveque2007finite}, finite element method~\citep{bathe2006finite}, finite volume method~\citep{eymard2000finite} or spectral methods~\citep{trefethen2000spectral}, followed by numerical solution of the resulting algebraic equations. While these methods have been successful in solving a wide range of PDEs, they often face the curse of dimensionality when parametric PDEs need to be solved in connection with optimization, optimal control, parameter identification, uncertainty quantification. The reduction of complexity for such classes of problems can be addressed with surrogate models using machine learning.

The main approaches in scientific machine learning are (i) using governing equations as loss functions with physics-informed neural networks~\citep{karniadakis2021physics, cai2021physics, eivazi2024physics, raissi2019physics}; (ii) learning mappings between infinite-dimensional function spaces with neural operators~\citep{li2020fourier, fanaskov2023spectral, lu2021learning,li2024geometry, tran2021factorized}; (iii) hybrid approaches where machine learning techniques are incorporated into classical simulations~\citep{brunton2022data, schnell2024stabilizing, hsieh2019learning, ingraham2018learning}.

These surrogate models have shown significant potential in solving parametric PDEs, but a critical aspect of their development remains the availability of comprehensive datasets for validation. The accuracy and reliability of these machine learning-based approaches are highly dependent on the quality and diversity of the data used to train and test them. Without such datasets, the performance and generalization ability of these models cannot be adequately assessed, and their applicability to real-world problems may be limited. As new techniques and methods emerge in the future, the need for robust and extensive datasets will only increase. It is therefore essential to develop approaches to the curation of high quality datasets that can support the development and validation of innovative approaches to solving complex problems in different scientific and engineering domains. 

Typically, scientific machine learning datasets have a large number of parametric PDEs~\citep{takamoto2022pdebench, luo2023cfdbench, hao2023pinnacle} that have a single example per PDE. With ConDiff (short for Contrast Diffusion) we focused on the idea of providing a large number of different realizations for a single problem - the diffusion equation. Currently, ConDiff consists of a diverse set of diffusion equations with $24$ realizations, which can be distinguished by complexity, and results in a total of $28800$ samples. We also propose an approach to generating complex coefficients for parametric PDEs that can address real-world problems with a measurable metric of the complexity of the dataset.

The ConDiff dataset is available on the Hugging Face Hub: \url{https://huggingface.co/datasets/condiff/ConDiff}. The code with ConDiff generation, usage, validation and requirements is available at: \url{https://github.com/condiff-dataset/ConDiff}. 


\section{ConDiff}
\label{sec:condiff}

\paragraph{Motivation}

Creating a comprehensive benchmark for classes of parametric PDEs is a particular challenge for the scientific machine learning community. The main challenges in creating a comprehensive dataset are: (i) computational complexity; (ii) storage complexity for the desired dimensions of the discretized PDE and parameter space; (iii) properties of the coefficients and solution functions; (iv) relation to real-world problems. The first and second reasons illustrate a technical bottleneck in the creation of the dataset and are mostly dependent on the hardware and efficiency of the numerical method used. Properties such as coefficient smoothness, discontinuity, spatial variation of the coefficients, variance of the parametric space significantly affect the complexity of the dataset and should be carefully chosen. The solution to parametric PDEs (i.e. the ground truth for the dataset) depends on a number of numerical aspects such as choice of mesh, discretization, numerical algorithm, boundary and initial conditions. Therefore, it is very important to consider every little detail regarding different numerical schemes, PDEs, boundary and initial conditions.

Existing benchmarks and datasets cover different aspects of scientific machine learning for different classes of PDEs and can be divided into several groups. PDEBench~\citep{takamoto2022pdebench}, PINNacle~\citep{hao2023pinnacle}, CFDBench~\citep{luo2023cfdbench} have a large number of PDEs with different boundary and initial conditions and different dimensionality and resolution. The best covered area is weather forecasting: SuperBench~\citep{ren2023superbench}, ClimSim~\citep{yu2024climsim}, DynaBench~\citep{dulny2023dynabench}, OceanBench~\citep{johnson2024oceanbench}, ChaosBench~\citep{nathaniel2024chaosbench}. There are also domain specific datasets with applications to Lagrangian mechanics LagrangeBench~\citep{toshev2024lagrangebench} and phase change phenomena BubbleML~\citep{hassan2023bubbleml}. Recently, the FlowBench~\citep{tali2024flowbench} dataset with complex geometries was introduced. Worth noting frameworks for differential simulations and general environments for PDEs in scientific machine learning: PDE Control Gym~\citep{bhan2024pde}, PDEArena~\citep{gupta2022towards}, DiffTaichi~\citep{hu2019difftaichi}, DeepXDE~\citep{lu2021deepxde} and $\Phi_{\text{Flow}}$~\citep{holl2020learning}.

While all of these datasets contribute significantly to the community, to the best of the authors' knowledge there is no dataset dedicated to the very important class of academic and real-world problems, the class of parametric PDEs with random coefficients. Typically, when a new model is proposed, authors test it with a set of equations with smooth coefficients~\citep{brandstetter2022message, nguyen2023neural, ripken2023multiscale, bryutkin2024hamlet}. Such coefficients do not allow important classes of industrial applications to be addressed. In section~\ref{sec:experiments} we show that increasing the heterogeneity and contrast in the coefficient function leads to increasing challenges in building accurate surrogate models.

\paragraph{Problem definition}

Existing benchmarks~\citep{takamoto2022pdebench, hao2023pinnacle, luo2023cfdbench} cover a set of PDEs, both steady-state and time-dependent, with different resolutions and time lengths. In our work, we approach the problem from the other side tacking a fixed parametric PDE and generating a comprehensive set of random coefficients for it. We consider a 2D steady-state diffusion equation:

\begin{equation}
    \label{eq:diffusion_equation}
    \begin{split}
    -&\nabla \cdot \big(k(x)\nabla u(x)\big) = f(x), ~\text{in}~\Omega\\
     &u(x)\Big|_{x\in \partial{\Omega}} = 0
    \end{split} \,\,\, .
\end{equation}

\begin{figure*}[!h]
\normalsize
\centering
    \begin{center}
        \makebox[\textwidth][c]{\includegraphics[width=1.0\textwidth]{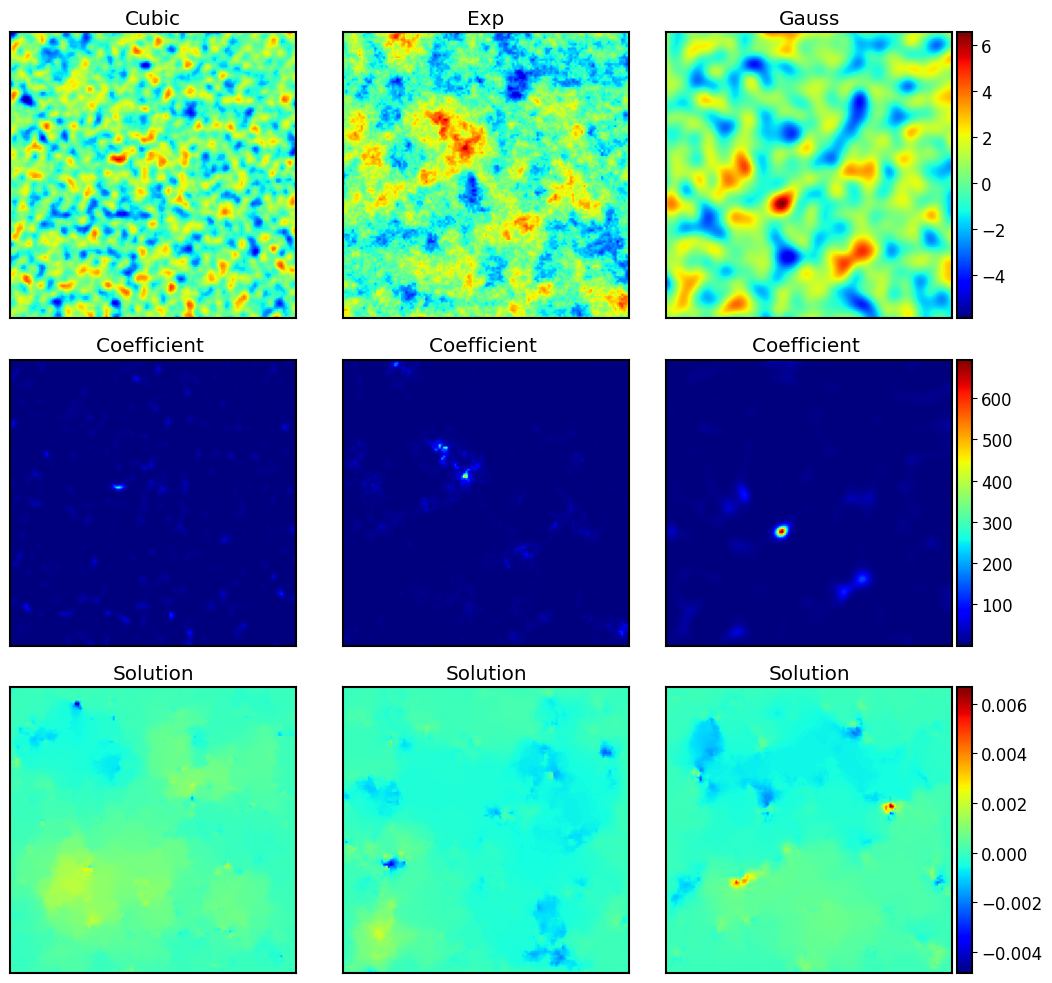}}%
    \end{center}
    \caption{Visualization of the GRF (top row), the coefficient $k(x)$ generated from this GRF (middle row) and the corresponding solution of the equation~(\ref{eq:diffusion_equation}) (bottom row) for a sampled PDEs with grid $128\times128$ and $\sigma^2=2.0$.}
    \label{fig:grf_coeff_sol}
    \vspace*{4pt}
\end{figure*}

Note that the equation~(\ref{eq:diffusion_equation}) models not only diffusion, but also steady-state Darcy flow in porous media, steady-state heat conduction, etc. To address certain real-world problems, we use the Gaussian Random Field (GRF) to generate the field $\phi(x)$ (Figure~\ref{fig:grf_coeff_sol}) with the following covariance models as functions of distance $d$:

\begin{itemize}
    \item Cubic:
    \begin{equation}
        \label{eq:cov_cubic}
        \text{Cov}(d) = 
            \begin{cases}
            \sigma^2 \Big( 1 - 7\big(\frac{d}{l}\big)^2 + \frac{35}{4}\big(\frac{d}{l}\big)^3 - \frac{7}{2}\big(\frac{d}{l}\big)^5 + \frac{3}{4}\big(\frac{d}{l}\big)^7 \Big)\, , & d < l \\
            0\, , & d \geq l
            \end{cases}\, .
    \end{equation}
    
    \item Exponential:
    \begin{equation}
        \label{eq:cov_exp}
        \text{Cov}(d) = \sigma^2 \exp{\Big(-\frac{d}{l}\Big)}\, .
    \end{equation}
    
    \item Gaussian:
    \begin{equation}
        \label{eq:cov_gauss}
        \text{Cov}(d) = \sigma^2 \exp{\Big(-\frac{d^2}{l^2}\Big)}\, .
    \end{equation}
\end{itemize}

The correlation length in each dataset is $l=0.05$ and the complexity of a resulting dataset is controlled by variance $\sigma^2$. The forcing term $f(x)$ is sampled from the standard normal distribution for each sampled PDE in each dataset. The resulting coefficient $k(x)$ is obtained with:

\begin{equation}
    \label{eq:diff_coeff}
    k(x) = \exp{\big(\phi(x)\big)} \, .
\end{equation}

We propose to measure the complexity of the generated GRF with the global contrast in the field $\phi(x)$:

\begin{equation}
    \label{eq:contrast}
    \text{contrast} = \exp\Big( \max \big(\phi(x)\big) - \min \big(\phi(x)\big) \Big) \, .
\end{equation}    

\paragraph{Complexity grows with variance}

By increasing the variance $\sigma^2$ one can obtain a higher contrast~(\ref{eq:contrast}) and thus a higher complexity of the PDE. This is a well-known phenomenon in applied numerical analysis and can be easily observed empirically. We illustrate this behaviour with the condition number $\kappa(A)$ of the matrices $A$ obtained with discretization of the equation~(\ref{eq:diffusion_equation}).

In the Table~\ref{table:ConDiff} one can observe that increasing $\sigma^2$ leads to a higher condition number $\kappa (A) = |\lambda_{\max}| \big/ |\lambda_{\min}|$ of the discretized differential operator~\citep{capizzano2003generalized}. The condition number is closely related to the performance of the numerical methods used to solve PDEs~\citep{benzi2005numerical, elman2014finite}. A high condition number indicates that small changes in the input can lead to large changes in the output, making the problem ill-conditioned. This is particularly important in PDEs, where small perturbations can significantly affect the solution. Also, if iterative methods are used to solve the discretized PDE, a larger condition number means a larger number of iterations for unpreconditioned and most of preconditioned iterative methods~\citep{saad2003iterative}.

\begin{figure*}[!h]
\normalsize
\centering
    \begin{center}
        \makebox[\textwidth][c]{\includegraphics[width=1.0\textwidth]{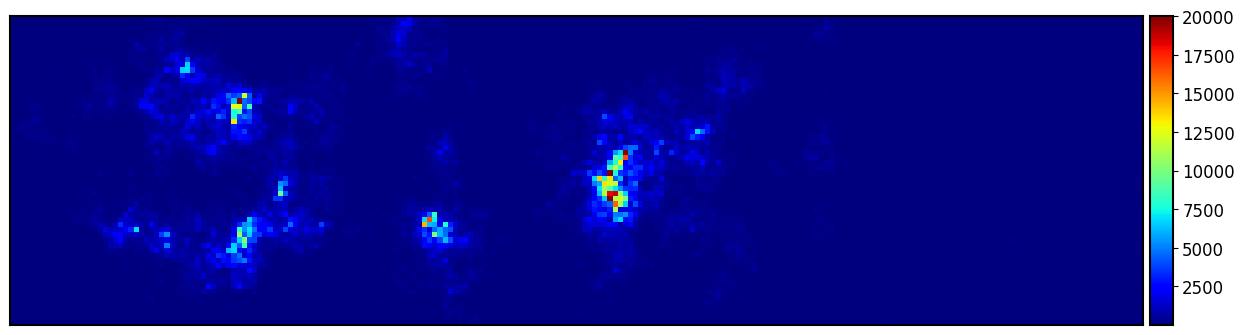}}%
    \end{center}
    \caption{Cross section of the $x-$permeability field along the $z$ axis over the SPE10 model 2 with $z=4$.}
    \label{fig:spe10_m2}
    \vspace*{4pt}
\end{figure*}

\paragraph{Connection to real-world problems}

All of the above reasoning is done with regard to the frequent occurrence of such tasks in real world~\citep{hashmi2014comprehensive, massimo2013physics, carr2016semi, oristaglio1984diffusion, muravleva2021multigrid}, including composite materials modeling, heat transfer, geophysical problems, fluid flow modeling. In Figure~\ref{fig:spe10_m2} one can see a cross section of the $x-$permeability field along the $z$ axis over the SPE10 model 2 benchmark~\citep{christie2001tenth}. The term permeability is used to denote the coefficients of the above equation when considering flow in porous media. This field is very similar to the ConDiff samples in Figure~\ref{fig:grf_coeff_sol}.

This benchmark is well known in the field of reservoir modelling and fluid flow in porous media. SPE10 model~2 poses a significant challenge for the tasks of uncertainty quantification, upscaling and multiphase fluid flow modelling. 

\begin{table}[!h]
    \caption{Summary of the ConDiff with min, mean and max values of the contrast~(\ref{eq:contrast}). $^{1}$Condition number $\kappa(A)$ is calculated for a single sampled discretized~(\ref{eq:diffusion_equation}).}
    \centering
    \begin{tabular}{@{}lccccr@{}}
        \\\toprule 
        Covariance & Variance & $\text{Min}~\text{contrast}$ & $\text{Mean} ~\text{contrast}$ & $\text{Max}~\text{contrast}$ & $\kappa^1(A)$ \\
        \midrule
        \multicolumn{6}{c}{$\text{Grid}~64\times64$}\\
        \midrule
        \multirow{4}{*}{$\text{Cubic}$} & $0.1$ & $7.0\cdot 10^{0}$ & $1.0\cdot 10^{1}$ & $1.5\cdot 10^{1}$ & $3.6\cdot 10^{3}$ \\ 
        & $0.4$ & $5.0\cdot 10^{1}$ & $9.6\cdot 10^{1}$ & $2.5\cdot 10^{2}$ & $7.3\cdot 10^{3}$ \\ 
        & $1.0$ & $6.0\cdot 10^{2}$ & $8.3\cdot 10^{2}$ & $1.0\cdot 10^{3}$ & $2.0\cdot 10^{4}$ \\ 
        & $2.0$ & $8.0\cdot 10^{4}$ & $8.9\cdot 10^{4}$ & $1.0\cdot 10^{5}$ & $1.8\cdot 10^{5}$ \\
        \midrule
        \multirow{4}{*}{$\text{Exp}$} & $0.1$ & $6.0\cdot 10^{0}$ & $9.0\cdot 10^{0}$ & $1.5\cdot 10^{1}$ & $4.3\cdot 10^{3}$ \\ 
        & $0.4$ & $5.0\cdot 10^{1}$ & $8.5\cdot 10^{1}$ & $2.3\cdot 10^{2}$ & $5.2\cdot 10^{3}$ \\ 
        & $1.0$ & $6.0\cdot 10^{2}$ & $7.9\cdot 10^{2}$ & $1.0\cdot 10^{3}$ & $1.7\cdot 10^{4}$ \\ 
        & $2.0$ & $8.0\cdot 10^{4}$ & $8.9\cdot 10^{4}$ & $1.0\cdot 10^{5}$ & $1.9\cdot 10^{5}$ \\ 
        \midrule
        \multirow{4}{*}{$\text{Gauss}$} & $0.1$ & $5.0\cdot 10^{0}$ & $8.0\cdot 10^{0}$ & $1.4\cdot 10^{1}$ & $4.1\cdot 10^{3}$ \\ 
        & 0$.4$ & $5.0\cdot 10^{1}$ & $7.5\cdot 10^{1}$ & $2.3\cdot 10^{2}$ & $8.1\cdot 10^{3}$ \\ 
        & $1.0$ & $6.0\cdot 10^{2}$ & $7.7\cdot 10^{2}$ & $1.0\cdot 10^{3}$ & $2.4\cdot 10^{4}$ \\  
        & $2.0$ & $8.0\cdot 10^{4}$ & $8.9\cdot 10^{4}$ & $1.0\cdot 10^{5}$ & $8.8\cdot 10^{5}$ \\ 
        \midrule
        \multicolumn{6}{c}{$\text{Grid}~128\times128$}\\
        \midrule
        \multirow{4}{*}{$\text{Cubic}$} & $0.1$ & $8.0\cdot 10^{0}$ & $1.1\cdot 10^{1}$ & $1.5\cdot 10^{1}$ & $1.6\cdot 10^{4}$ \\ 
        & $0.4$ & $5.5\cdot 10^{1}$ & $1.3\cdot 10^{2}$ & $2.5\cdot 10^{2}$ & $3.8\cdot 10^{4}$ \\ 
        & $1.0$ & $6.0\cdot 10^{2}$ & $8.8\cdot 10^{2}$ & $1.0\cdot 10^{3}$ & $1.0\cdot 10^{5}$ \\ 
        & $2.0$ & $8.0\cdot 10^{4}$ & $8.9\cdot 10^{4}$ & $1.0\cdot 10^{5}$ & $1.2\cdot 10^{6}$ \\
        \midrule
        \multirow{4}{*}{$\text{Exp}$} & $0.1$ & $6.0\cdot 10^{0}$ & $1.0\cdot 10^{1}$ & $1.5\cdot 10^{1}$ & $1.7\cdot 10^{4}$ \\ 
        & $0.4$ & $5.1\cdot 10^{1}$ & $1.1\cdot 10^{2}$ & $2.5\cdot 10^{2}$ & $3.3\cdot 10^{4}$ \\ 
        & $1.0$ & $6.0\cdot 10^{2}$ & $8.3\cdot 10^{2}$ & $1.0\cdot 10^{3}$ & $9.7\cdot 10^{4}$ \\ 
        & $2.0$ & $8.0\cdot 10^{4}$ & $8.9\cdot 10^{4}$ & $1.0\cdot 10^{5}$ & $6.3\cdot 10^{5}$ \\
        \midrule
        \multirow{4}{*}{$\text{Gauss}$} & $0.1$ & $5.0\cdot 10^{0}$ & $8.0\cdot 10^{0}$ & $1.4\cdot 10^{1}$ & $1.8\cdot 10^{4}$ \\ 
        & $0.4$ & $5.0\cdot 10^{1}$ & $7.8\cdot 10^{1}$ & $2.5\cdot 10^{2}$ & $7.2\cdot 10^{4}$ \\ 
        & $1.0$ & $6.0\cdot 10^{2}$ & $7.7\cdot 10^{2}$ & $1.0\cdot 10^{3}$ & $1.6\cdot 10^{5}$ \\ 
        & $2.0$ & $8.0\cdot 10^{4}$ & $8.9\cdot 10^{4}$ & $1.0\cdot 10^{5}$ & $1.5\cdot 10^{6}$ \\
        \bottomrule
    \end{tabular}
    \label{table:ConDiff}
\end{table}

\paragraph{Dataset description} 

To generate the fields $\phi(x)$ we use the highly efficient \texttt{parafields} library\footnote{https://github.com/parafields/parafields} with C++ backend. We use covariance models from $\{\text{cubic}, \text{exponential}, \text{Gaussian}\}$ with 4 variance values from $\{0.1, 0.4, 1.0, 2.0\}$. We use the forcing term $f(x) \sim \mathcal{N}(0, 1)$. The standard normal force function is chosen to be more complex than a constant forcing term, but not too complex to distract from the complex coefficients, which is the focus of ConDiff. A Dirichlet boundary condition is set for each coefficient realization since boundary conditions do not contribute significantly to the resulting complexity~\citep{capizzano2003generalized}. The ground truth solution is obtained using cell-centered second-order finite volume method. The coefficients are in the center of cells, the values are in the nodes.

For each parameter set, we generate $1000$ training and $200$ test realizations of the diffusion equation~(\ref{eq:diffusion_equation}) on $64\times64$ and $128\times128$ grids. We provide the train-test split in the ConDiff for fair comparison in future research papers. Note that datasets with the same field parameters but different grid sizes are generated independently and do not represent the same field. The fixed geometry of ConDiff allows PDEs with different fields $\phi(x)$ to be compared without fear that different geometries will interfere with a fair comparison across different coefficient functions. To control the complexity of the generated PDEs realizations, we set contrast bounds during generation as follows: 

\begin{itemize}
    \item $\sigma^2 = 0.1,~\text{contrast} \in [5, 15]$,
    \item $\sigma^2 = 0.4,~\text{contrast} \in [50, 250]$,
    \item $\sigma^2 = 1.0,~\text{contrast} \in [6\cdot10^2, 10^3]$,
    \item $\sigma^2 = 2.0,~\text{contrast} \in [8\cdot10^4, 10^5]$.
\end{itemize}

\begin{figure*}[!h]
\normalsize
\centering
    \begin{center}
        \makebox[\textwidth][c]{\includegraphics[width=1.1\textwidth]{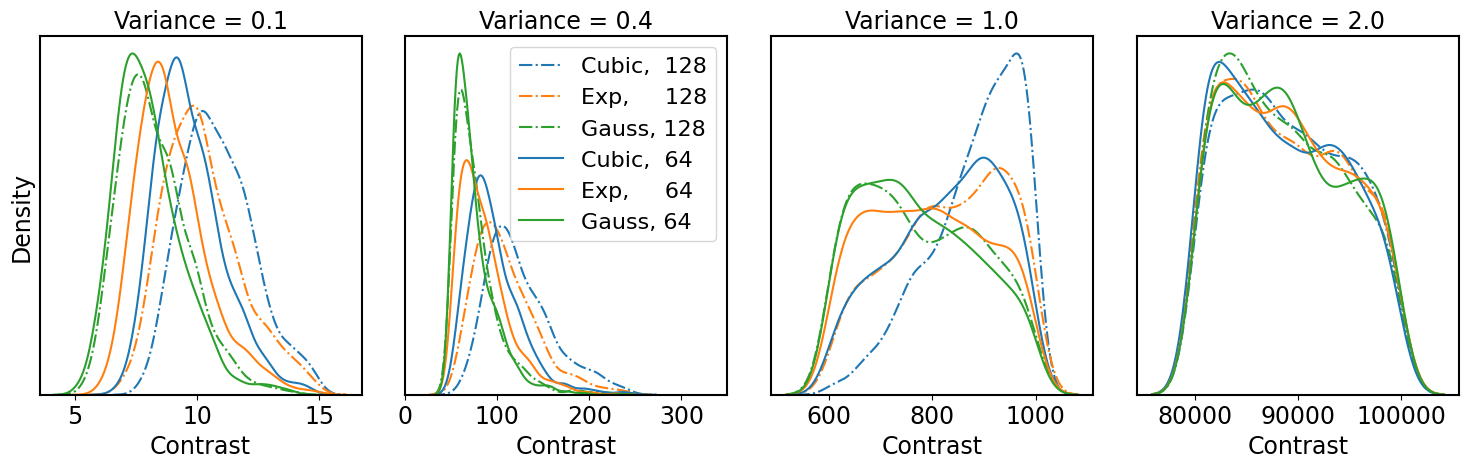}}%
    \end{center}
    \caption{GRF contrast distribution for PDEs from Table~\ref{table:ConDiff}.}
    \label{fig:contrast_distrib}
    \vspace*{4pt}
\end{figure*}

In total, ConDiff consists of 24 PDEs with different GRFs and grid sizes. Table~\ref{table:ConDiff} summarizes the properties of ConDiff. Figure~\ref{fig:contrast_distrib} illustrates the contrast distributions. Coming back to the permeability cross section of SPE10 model~2 (Figure~\ref{fig:spe10_m2}), it has $\text{contrast}=2.5\cdot 10^{6}$ according to~(\ref{eq:contrast}). We want to emphasize that although the most complex coefficient of ConDiff is smaller by an order of magnitude compared to the cross section of SPE10 model~2, our experiments show that this coefficient is too complex for the chosen models to predict well.

\section{Experiments}
\label{sec:experiments}

\paragraph{Models} We do not attempt to benchmark every scientific machine learning surrogate model on the ConDiff. Since the ConDiff consists of triplets $\big( k(x), f(x), u(x) \big)$, its primary use is to validate different architectures of neural operators. Therefore, we have selected the following list of models to validate on the ConDiff: Spectral Neural Operator (SNO)~\citep{fanaskov2023spectral}, Factorized Fourier Neural Operator (F-FNO)~\citep{tran2021factorized}, Dilated ResNet (DilResNet)~\citep{yu2017dilated} and U-Net~\citep{ronneberger2015u}. Neural operators FNO and SNO are both types of neural networks designed to learn mappings between function spaces, in particular to solve PDEs. Neural operators are designed to be universal approximators of continuous operators acting between Banach spaces and to be discretization invariant, meaning that they can handle different discretizations of the underlying function spaces without requiring changes to the model. DilResNet and U-Net are classical neural network models originating from the field of computer vision (CV). Both models have shown their applicability beyond CV and have been used extensively for modeling physical phenomena~\citep{stachenfeld2021learned, ma2021physics}. More details about the models used can be found in the Appendix~\ref{app:models}.

\paragraph{Experiment environment} For training neural networks we use frameworks from the JAX~\citep{jax2018github} ecosystem: Equinox~\citep{kidger2021equinox} and Optax~\citep{deepmind2020jax}. The loss function used is the relative $L_2$ loss:

\begin{equation}
    \label{eq:L2 error}
    L_2 = \frac{1}{N} \sum_{i=1}^{N} \frac{\Vert \hat{y_i} - y_i \Vert_2}{\Vert y_i\Vert_2}~.
\end{equation}

Training samples for the models are the values of the coefficient function $k(x)$ and the forcing term $f(x)$ in the grid cells. Targets are the values of the solution function $u(x)$ in the grid cells. We also use~(\ref{eq:L2 error}) as a primary performance metric, assessing the quality of the models' predictions, and report averaged values over the test set with standard deviation.

For all the problems we train for $400$ epochs for $\text{grid}=64$ and for $500$ epochs for $\text{grid}=128$. We use the AdamW optimizer with an initial learning rate equals to $10^{-3}$ and a weight decay equals to $10^{-2}$. We use a learning rate schedule that halves the learning rate every $50$ epochs. Each PDE realization has a dataset size of $1000$ training samples and $200$ test samples. We use a single GPU Nvidia Tesla V100 16Gb for training on $\text{grid}=64$ and a single GPU Nvidia A40 48Gb for training on $\text{grid}=128$.

\begin{table}[!h]
    \centering
    \caption{Results for Poisson equation.}
    \label{table:results_poisson_grids_64_128}
    \centering
    \begin{tabular}{@{}ccccccccc@{}}
        \\\toprule
        Grid & \multicolumn{2}{c}{$\text{SNO}$} & \multicolumn{2}{c}{$\text{F-FNO}$} & \multicolumn{2}{c}{$\text{DilResNet}$} & \multicolumn{2}{c}{$\text{U-Net}$} \\\midrule
        64 & \multicolumn{2}{c}{$0.056 \pm 0.018$} & \multicolumn{2}{c}{$0.027 \pm 0.008$} & \multicolumn{2}{c}{$0.018 \pm 0.005$} & \multicolumn{2}{c}{$0.020 \pm 0.007$} \\[0.1cm]
        128 & \multicolumn{2}{c}{$0.073 \pm 0.021$} & \multicolumn{2}{c}{$0.047 \pm 0.013$} & \multicolumn{2}{c}{$0.063 \pm 0.016$} & \multicolumn{2}{c}{$0.267 \pm 0.049$} \\
        \bottomrule 
    \end{tabular}
\end{table}
\begin{table}[!h]
    \caption{Performance comparison of the models on the PDEs with the $64\times64$ grid from ConDiff.}
    \label{table:results_grid_64}
    \centering
    \begin{tabular}{@{}lccccccccc@{}}
        \\\toprule
        Covariance & Variance & \multicolumn{2}{c}{$\text{SNO}$} & \multicolumn{2}{c}{$\text{F-FNO}$} & \multicolumn{2}{c}{$\text{DilResNet}$}  & \multicolumn{2}{c}{$\text{U-Net}$} \\\cmidrule(r){1-2}\cmidrule(r){3-4} \cmidrule(r){5-6} \cmidrule(r){7-8} \cmidrule(r){9-10} 
        \multirow{4}{*}{\centering \text{Cubic}} & $0.1$ & \multicolumn{2}{c}{$0.09 \pm 0.02$} & \multicolumn{2}{c}{$0.07 \pm 0.02$} & \multicolumn{2}{c}{$0.07 \pm 0.02$} & \multicolumn{2}{c}{$0.08 \pm 0.02$}\\
        & $0.4$ & \multicolumn{2}{c}{$0.15 \pm 0.04$} & \multicolumn{2}{c}{$0.14 \pm 0.03$} & \multicolumn{2}{c}{$0.14 \pm 0.03$} & \multicolumn{2}{c}{$0.17 \pm 0.04$} \\
        & $1.0$ & \multicolumn{2}{c}{$0.23 \pm 0.06$} & \multicolumn{2}{c}{$0.22 \pm 0.06$} & \multicolumn{2}{c}{$0.22 \pm 0.06$} & \multicolumn{2}{c}{$0.24 \pm 0.06$} \\
        & $2.0$ & \multicolumn{2}{c}{$0.35 \pm 0.10$} & \multicolumn{2}{c}{$0.34 \pm 0.09$} & \multicolumn{2}{c}{$0.35 \pm 0.10$} & \multicolumn{2}{c}{$0.42 \pm 0.10$} \\
        \midrule
        \multirow{4}{*}{\centering \text{Exp}} & $0.1$ & \multicolumn{2}{c}{$0.12 \pm 0.03$} & \multicolumn{2}{c}{$0.11 \pm 0.03$} & \multicolumn{2}{c}{$0.11 \pm 0.03$} & \multicolumn{2}{c}{$0.12 \pm 0.04$}\\
        & $0.4$ & \multicolumn{2}{c}{$0.21 \pm 0.06$} & \multicolumn{2}{c}{$0.21 \pm 0.06$} & \multicolumn{2}{c}{$0.20 \pm 0.06$} & \multicolumn{2}{c}{$0.26 \pm 0.07$} \\
        & $1.0$ & \multicolumn{2}{c}{$0.33 \pm 0.09$} & \multicolumn{2}{c}{$0.34 \pm 0.09$} & \multicolumn{2}{c}{$0.36 \pm 0.09$} & \multicolumn{2}{c}{$0.35 \pm 0.09$} \\
        & $2.0$ & \multicolumn{2}{c}{$0.59 \pm 0.14$} & \multicolumn{2}{c}{$0.58 \pm 0.14$} & \multicolumn{2}{c}{$0.60 \pm 0.13$} & \multicolumn{2}{c}{$0.64 \pm 0.13$} \\
        \midrule
        \multirow{4}{*}{\centering \text{Gauss}} & $0.1$ & \multicolumn{2}{c}{$0.12 \pm 0.04$} & \multicolumn{2}{c}{$0.11 \pm 0.04$} & \multicolumn{2}{c}{$0.11 \pm 0.03$} & \multicolumn{2}{c}{$0.12 \pm 0.03$}\\
        & $0.4$ & \multicolumn{2}{c}{$0.23 \pm 0.06$} & \multicolumn{2}{c}{$0.22 \pm 0.06$} & \multicolumn{2}{c}{$0.21 \pm 0.06$} & \multicolumn{2}{c}{$0.25 \pm 0.06$} \\
        & $1.0$ & \multicolumn{2}{c}{$0.38 \pm 0.08$} & \multicolumn{2}{c}{$0.37 \pm 0.09$} & \multicolumn{2}{c}{$0.38 \pm 0.09$} & \multicolumn{2}{c}{$0.39 \pm 0.09$} \\
        & $2.0$ & \multicolumn{2}{c}{$0.66 \pm 0.14$} & \multicolumn{2}{c}{$0.65 \pm 0.14$} & \multicolumn{2}{c}{$0.66 \pm 0.13$} & \multicolumn{2}{c}{$0.72 \pm 0.24$} \\
        \bottomrule 
    \end{tabular}
\end{table}
\begin{table}[!h]
    \caption{Performance comparison of SNO and F-FNO on the PDEs with the $128\times128$ grid from ConDiff.}
    \label{table:results_grid_128}
    \centering
    \begin{tabular}{@{}lccccc@{}}
        \\\toprule
        Covariance & Variance & \multicolumn{2}{c}{$\text{SNO}$} & \multicolumn{2}{c}{$\text{F-FNO}$} \\\cmidrule(r){1-2}\cmidrule(r){3-4} \cmidrule(r){5-6}
        \multirow{4}{*}{\centering \text{Cubic}} & $0.1$ & \multicolumn{2}{c}{$0.09 \pm 0.03$} & \multicolumn{2}{c}{$0.08 \pm 0.02$} \\
        & $0.4$ & \multicolumn{2}{c}{$0.15 \pm 0.04$} & \multicolumn{2}{c}{$0.14 \pm 0.04$} \\
        & $1.0$ & \multicolumn{2}{c}{$0.23 \pm 0.06$} & \multicolumn{2}{c}{$0.22 \pm 0.06$} \\
        & $2.0$ & \multicolumn{2}{c}{$0.36 \pm 0.11$} & \multicolumn{2}{c}{$0.36 \pm 0.10$} \\
        \midrule
        \multirow{4}{*}{\centering \text{Exp}} & $0.1$ & \multicolumn{2}{c}{$0.13 \pm 0.03$} & \multicolumn{2}{c}{$0.12 \pm 0.03$} \\
        & $0.4$ & \multicolumn{2}{c}{$0.21 \pm 0.07$} & \multicolumn{2}{c}{$0.21 \pm 0.06$}  \\
        & $1.0$ & \multicolumn{2}{c}{$0.33 \pm 0.09$} & \multicolumn{2}{c}{$0.33 \pm 0.08$}  \\
        & $2.0$ & \multicolumn{2}{c}{$0.58 \pm 0.15$} & \multicolumn{2}{c}{$0.57 \pm 0.13$} \\
        \midrule
        \multirow{4}{*}{\centering \text{Gauss}} & $0.1$ & \multicolumn{2}{c}{$0.13 \pm 0.04$} & \multicolumn{2}{c}{$0.12 \pm 0.03$} \\
        & $0.4$ & \multicolumn{2}{c}{$0.23 \pm 0.06$} & \multicolumn{2}{c}{$0.23 \pm 0.06$}  \\
        & $1.0$ & \multicolumn{2}{c}{$0.37 \pm 0.10$} & \multicolumn{2}{c}{$0.37 \pm 0.10$}  \\
        & $2.0$ & \multicolumn{2}{c}{$0.68 \pm 0.13$} & \multicolumn{2}{c}{$0.66 \pm 0.13$}  \\
        \bottomrule 
    \end{tabular}
\end{table}

\paragraph{Validation on ConDiff} We start the experiments with the Poisson equation and consider it as a special case of~(\ref{eq:diffusion_equation}) with $k(x) = 1$ and $\text{contrast} = 1$. All models achieve an accuracy of the order of $10^{-2}$ (Table~\ref{table:results_poisson_grids_64_128}). Increasing the grid size leads to moderate increases in error, except for the U-Net for which the error increases by an order of magnitude.

The diffusion equation for grid $64$ (Table~\ref{table:results_grid_64}) with covariances~(\ref{eq:cov_cubic}),~(\ref{eq:cov_exp}) and~(\ref{eq:cov_gauss}) are more challenging for the models. While the performance on the diffusion equation with cubic covariance with $\sigma^2 = 0.1$ is comparable to the performance on the Poisson equation, the error  on the diffusion equation with exponential and Gaussian covariances is already an order of magnitude higher. Increasing $\sigma^2$ leads to worse performance of each model on each PDE. The most complex PDE is the one generated with the Gaussian covariance model in GRF, which is also consistent with the condition number estimation in Table~\ref{table:ConDiff}. Interestingly, the performance of FNO and SNO models on PDEs with grid $128$ is not much different from PDEs on grid $64$ (Table~\ref{table:results_grid_128}).

\begin{table}[!h]
    \caption{Generalization of the models to unseen PDEs with different GRF covariance model with $64\times64$ grid and $\sigma^2=0.1$.}
    \label{table:transfer_cov_var0.1}
    \centering
    \begin{tabular}{@{}lcccccc@{}}
        \\\toprule
        & \multicolumn{3}{c}{$\text{SNO}$} & \multicolumn{3}{c}{$\text{F-FNO}$}\\\cmidrule(r){2-4} \cmidrule(r){5-7}
        $\text{Train} \big\backslash \text{Test}$ &  \text{Cubic} & \text{Exp} & \text{Gauss} & \text{Cubic} & \text{Exp} & \text{Gauss} \\
        \midrule
        \text{Cubic} & $0.09 \pm 0.02$ & $0.12 \pm 0.04$ & $0.12 \pm 0.03$ & $0.07 \pm 0.02$ & $0.11 \pm 0.03$ & $0.11 \pm 0.03$   \\\cmidrule(r){2-4}\cmidrule(r){5-7}
        \text{Exp} & $0.09 \pm 0.03$ & $0.12 \pm 0.03$ & $0.12 \pm 0.04$ & $0.08 \pm 0.03$ & $0.11 \pm 0.03$ & $0.11 \pm 0.04$   \\\cmidrule(r){2-4}\cmidrule(r){5-7}
        \text{Gauss} & $0.09 \pm 0.03$ & $0.12 \pm 0.03$ & $0.12 \pm 0.03$ & $0.08 \pm 0.02$ & $0.11 \pm 0.03$ & $0.11 \pm 0.04$   \\
        \midrule
        & \multicolumn{3}{c}{$\text{DilResNet}$} & \multicolumn{3}{c}{$\text{U-Net}$}\\\cmidrule(r){2-4} \cmidrule(r){5-7}
        &  \text{Cubic} & \text{Exp} & \text{Gauss} & \text{Cubic} & \text{Exp} & \text{Gauss} \\
        \midrule
        \text{Cubic} & $0.07 \pm 0.02$ & $0.11 \pm 0.04$ & $0.11 \pm 0.03$ & $0.08 \pm 0.02$ & $0.12 \pm 0.03$ & $0.12 \pm 0.03$   \\\cmidrule(r){2-4}\cmidrule(r){5-7}
        \text{Exp} & $0.07 \pm 0.02$ & $0.11 \pm 0.03$ & $0.11 \pm 0.03$ & $0.08 \pm 0.03$ & $0.11 \pm 0.03$ & $0.11 \pm 0.04$   \\\cmidrule(r){2-4}\cmidrule(r){5-7}
        \text{Gauss} & $0.17 \pm 0.06$ & $0.25 \pm 0.09$ & $0.11 \pm 0.04$ & $0.08 \pm 0.02$ & $0.12 \pm 0.04$ & $0.12 \pm 0.03$   \\
        \bottomrule 
    \end{tabular}
\end{table}
\begin{table}[!h]
    \caption{Generalization of the models to unseen PDEs with different GRF covariance model with $64\times64$ grid and $\sigma^2=0.4$.}
    \label{table:transfer_cov_var0.4}
    \centering
    \begin{tabular}{@{}lcccccc@{}}
        \\\toprule
        & \multicolumn{3}{c}{$\text{SNO}$} & \multicolumn{3}{c}{$\text{F-FNO}$}\\\cmidrule(r){2-4} \cmidrule(r){5-7}
        $\text{Train} \big\backslash \text{Test}$ &  \text{Cubic} & \text{Exp} & \text{Gauss} & \text{Cubic} & \text{Exp} & \text{Gauss} \\
        \midrule
        \text{Cubic} & $0.15 \pm 0.04$ & $0.22 \pm 0.06$ & $0.22 \pm 0.07$ & $0.14 \pm 0.03$ & $0.21 \pm 0.06$ & $0.21 \pm 0.07$   \\\cmidrule(r){2-4}\cmidrule(r){5-7}
        \text{Exp} & $0.18 \pm 0.05$ & $0.21 \pm 0.06$ & $0.22 \pm 0.06$ & $0.15 \pm 0.04$ & $0.21 \pm 0.06$ & $0.22 \pm 0.07$   \\\cmidrule(r){2-4}\cmidrule(r){5-7}
        \text{Gauss} & $0.17 \pm 0.05$ & $0.22 \pm 0.06$ & $0.23 \pm 0.07$ & $0.15 \pm 0.04$ & $0.21 \pm 0.07$ & $0.22 \pm 0.06$   \\
        \midrule
        & \multicolumn{3}{c}{$\text{DilResNet}$} & \multicolumn{3}{c}{$\text{U-Net}$}\\\cmidrule(r){2-4} \cmidrule(r){5-7}
        &  \text{Cubic} & \text{Exp} & \text{Gauss} & \text{Cubic} & \text{Exp} & \text{Gauss} \\
        \midrule
        \text{Cubic} & $0.14 \pm 0.04$ & $0.23 \pm 0.07$ & $0.23 \pm 0.07$ & $0.17 \pm 0.06$ & $0.24 \pm 0.07$ & $0.24 \pm 0.07$   \\\cmidrule(r){2-4}\cmidrule(r){5-7}
        \text{Exp} & $0.14 \pm 0.04$ & $0.20 \pm 0.06$ & $0.22 \pm 0.06$ & $0.23 \pm 0.08$ & $0.26 \pm 0.07$ & $0.27 \pm 0.08$   \\\cmidrule(r){2-4}\cmidrule(r){5-7}
        \text{Gauss} & $0.30 \pm 0.10$ & $0.24 \pm 0.07$ & $0.21 \pm 0.06$ & $0.21 \pm 0.06$ & $0.27 \pm 0.08$ & $0.26 \pm 0.07$   \\
        \bottomrule 
    \end{tabular}
\end{table}
\begin{table}[!h]
    \caption{Generalization of the models to unseen PDEs with different GRF covariance model with $64\times64$ grid and $\sigma^2=1.0$.}
    \label{table:transfer_cov_var1.0}
    \centering
    \begin{tabular}{@{}lcccccc@{}}
        \\\toprule
        & \multicolumn{3}{c}{$\text{SNO}$} & \multicolumn{3}{c}{$\text{F-FNO}$}\\\cmidrule(r){2-4} \cmidrule(r){5-7}
        $\text{Train} \big\backslash \text{Test}$ &  \text{Cubic} & \text{Exp} & \text{Gauss} & \text{Cubic} & \text{Exp} & \text{Gauss} \\
        \midrule
        \text{Cubic} & $0.23 \pm 0.06$ & $0.35 \pm 0.09$ & $0.39 \pm 0.09$ & $0.22 \pm 0.06$ & $0.34 \pm 0.09$ & $0.37 \pm 0.09$   \\\cmidrule(r){2-4}\cmidrule(r){5-7}
        \text{Exp} & $0.25 \pm 0.06$ & $0.33 \pm 0.09$ & $0.38 \pm 0.09$ & $0.24 \pm 0.06$ & $0.34 \pm 0.09$ & $0.38 \pm 0.09$   \\\cmidrule(r){2-4}\cmidrule(r){5-7}
        \text{Gauss} & $0.24 \pm 0.07$ & $0.35 \pm 0.09$ & $0.38 \pm 0.08$ & $0.24 \pm 0.06$ & $0.35 \pm 0.09$ & $0.37 \pm 0.09$   \\
        \midrule
        & \multicolumn{3}{c}{$\text{DilResNet}$} & \multicolumn{3}{c}{$\text{U-Net}$}\\\cmidrule(r){2-4} \cmidrule(r){5-7}
        &  \text{Cubic} & \text{Exp} & \text{Gauss} & \text{Cubic} & \text{Exp} & \text{Gauss} \\
        \midrule
        \text{Cubic} & $0.22 \pm 0.06$ & $0.35 \pm 0.09$ & $0.38 \pm 0.09$ & $0.24 \pm 0.06$ & $0.36 \pm 0.09$ & $0.38 \pm 0.08$   \\\cmidrule(r){2-4}\cmidrule(r){5-7}
        \text{Exp} & $0.25 \pm 0.07$ & $0.36 \pm 0.09$ & $0.38 \pm 0.10$ & $0.25 \pm 0.07$ & $0.35 \pm 0.09$ & $0.38 \pm 0.10$   \\\cmidrule(r){2-4}\cmidrule(r){5-7}
        \text{Gauss} & $0.57 \pm 0.22$ & $0.59 \pm 0.22$ & $0.38 \pm 0.09$ & $0.27 \pm 0.07$ & $0.36 \pm 0.11$ & $0.39 \pm 0.09$   \\
        \bottomrule 
    \end{tabular}
\end{table}
\begin{table}[!h]
    \caption{Generalization of the models to unseen PDEs with different GRF covariance model with $64\times64$ grid and $\sigma^2=2.0$.}
    \label{table:transfer_cov_var2.0}
    \centering
    \begin{tabular}{@{}lcccccc@{}}
        \\\toprule
        & \multicolumn{3}{c}{$\text{SNO}$} & \multicolumn{3}{c}{$\text{F-FNO}$}\\\cmidrule(r){2-4} \cmidrule(r){5-7}
        $\text{Train} \big\backslash \text{Test}$ &  \text{Cubic} & \text{Exp} & \text{Gauss} & \text{Cubic} & \text{Exp} & \text{Gauss} \\
        \midrule
        \text{Cubic} & $0.35 \pm 0.10$ & $0.60 \pm 0.14$ & $0.70 \pm 0.26$ & $0.34 \pm 0.09$ & $0.61 \pm 0.14$ & $0.67 \pm 0.19$   \\\cmidrule(r){2-4}\cmidrule(r){5-7}
        \text{Exp} & $0.39 \pm 0.11$ & $0.59 \pm 0.14$ & $0.69 \pm 0.24$ & $0.39 \pm 0.11$ & $0.58 \pm 0.14$ & $0.66 \pm 0.15$   \\\cmidrule(r){2-4}\cmidrule(r){5-7}
        \text{Gauss} & $0.40 \pm 0.11$ & $0.60 \pm 0.13$ & $0.66 \pm 0.14$ & $0.37 \pm 0.11$ & $0.60 \pm 0.13$ & $0.65 \pm 0.14$   \\
        \midrule
        & \multicolumn{3}{c}{$\text{DilResNet}$} & \multicolumn{3}{c}{$\text{U-Net}$}\\\cmidrule(r){2-4} \cmidrule(r){5-7}
        &  \text{Cubic} & \text{Exp} & \text{Gauss} & \text{Cubic} & \text{Exp} & \text{Gauss} \\
        \midrule
        \text{Cubic} & $0.35 \pm 0.10$ & $0.61 \pm 0.14$ & $0.66 \pm 0.15$ & $0.42 \pm 0.10$ & $0.65 \pm 0.14$ & $0.68 \pm 0.14$   \\\cmidrule(r){2-4}\cmidrule(r){5-7}
        \text{Exp} & $0.41 \pm 0.10$ & $0.60 \pm 0.13$ & $0.66 \pm 0.17$ & $0.53 \pm 0.18$ & $0.64 \pm 0.13$ & $0.72 \pm 0.16$   \\\cmidrule(r){2-4}\cmidrule(r){5-7}
        \text{Gauss} & $0.72 \pm 0.50$ & $0.68 \pm 0.20$ & $0.66 \pm 0.13$ & $0.66 \pm 0.40$ & $0.69 \pm 0.16$ & $0.72 \pm 0.24$   \\
        \bottomrule 
    \end{tabular}
\end{table}

\paragraph{Transfer between parametric spaces} Ideally, the surrogate model should handle transfers between different underlying parametric spaces of PDEs without loss of quality. In Tables~\ref{table:transfer_cov_var0.1}, \ref{table:transfer_cov_var0.4}, \ref{table:transfer_cov_var1.0}, \ref{table:transfer_cov_var2.0} show that in most experiments the error increases when training on cubic GRF and inferencing on exponential and Gaussian GRF. Conversely, the error decreases when training on Gaussian GRF and inferencing on cubic GRF. 

\section{Discussion}
\label{sec:conclusion}

We propose a novel dataset for the field of neural solving of parametric PDEs. The unique feature of the dataset is discontinuous coefficients with high contrast for parametric PDEs from different distributions. By designing the coefficients in this way, we achieve a high complexity of the generated PDEs, which also illustrates real-world problems. The proposed complexity function allows to distinguish between the generated PDEs. We also provide code to generate new data based on the approach used in this paper. Furthermore, we validate a number of surrogate models on the ConDiff to illustrate its usefulness in the field of scientific machine learning.

The practical use of ConDiff is straightforward: it should be used for novel deep learning models and approaches for modeling solution of parametric PDEs from their coefficients. Ultimately, novel deep learning models should exhibit machine-precision prediction quality and not degrade with increasing contrast.

It should be noted that the problems considered in this paper belong to the class of stochastic PDEs. The equation~(\ref{eq:diffusion_equation}) has to be solved for a very large number of sampled coefficients when Monte Carlo or other methods are used to solve the stochastic PDEs. The surrogate models can help to significantly reduce the computational burden, so embedding the surrogate models tested on ConDiff into a Monte Carlo or similar stochastic PDEs solver is a reasonable next step.

\section{Limitations}
\label{sec:limitations}

Limitations of the proposed dataset are:

\begin{itemize}
    \item[1.] For practical numerical analysis, ConDiff is generated with small and moderate variances. The case of large variances has to be studied separately.
    \item[2.] A linear elliptic parametric PDE is the basis of ConDiff, so other high contrast datasets are needed to test surrogate models for hyperbolic PDEs, nonlinear problems, etc.
    \item[3.] ConDiff is generated on a regular rectangular grid. Other meshes and geometries may be required as an evolution of ConDiff. This may require more complex computational methods to obtain the ground truth solution.
    \item[4.] The forcing term $f(x)$ is sampled from the standard normal distributions. While in this paper we focus on the complexity arising from discontinuous coefficients with high contrast, the right-hand side of a PDE can also significantly affect the complexity of the solving PDE. The case of complex forcing terms has to be studied separately.
\end{itemize}


\bibliography{iclr2025_conference}

\begin{thebibliography}{56}
\providecommand{\natexlab}[1]{#1}
\providecommand{\url}[1]{\texttt{#1}}
\expandafter\ifx\csname urlstyle\endcsname\relax
  \providecommand{\doi}[1]{doi: #1}\else
  \providecommand{\doi}{doi: \begingroup \urlstyle{rm}\Url}\fi

\bibitem[Bar-Sinai et~al.(2019)Bar-Sinai, Hoyer, Hickey, and Brenner]{bar2019learning}
Yohai Bar-Sinai, Stephan Hoyer, Jason Hickey, and Michael~P Brenner.
\newblock Learning data-driven discretizations for partial differential equations.
\newblock \emph{Proceedings of the National Academy of Sciences}, 116\penalty0 (31):\penalty0 15344--15349, 2019.

\bibitem[Bathe(2006)]{bathe2006finite}
Klaus-J{\"u}rgen Bathe.
\newblock \emph{Finite element procedures}.
\newblock Klaus-Jurgen Bathe, 2006.

\bibitem[Benzi et~al.(2005)Benzi, Golub, and Liesen]{benzi2005numerical}
Michele Benzi, Gene~H Golub, and J{\"o}rg Liesen.
\newblock Numerical solution of saddle point problems.
\newblock \emph{Acta numerica}, 14:\penalty0 1--137, 2005.

\bibitem[Bhan et~al.(2024)Bhan, Bian, Krstic, and Shi]{bhan2024pde}
Luke Bhan, Yuexin Bian, Miroslav Krstic, and Yuanyuan Shi.
\newblock Pde control gym: A benchmark for data-driven boundary control of partial differential equations.
\newblock \emph{arXiv preprint arXiv:2405.11401}, 2024.

\bibitem[Bradbury et~al.(2018)Bradbury, Frostig, Hawkins, Johnson, Leary, Maclaurin, Necula, Paszke, Vander{P}las, Wanderman-{M}ilne, and Zhang]{jax2018github}
James Bradbury, Roy Frostig, Peter Hawkins, Matthew~James Johnson, Chris Leary, Dougal Maclaurin, George Necula, Adam Paszke, Jake Vander{P}las, Skye Wanderman-{M}ilne, and Qiao Zhang.
\newblock {JAX}: composable transformations of {P}ython+{N}um{P}y programs, 2018.
\newblock URL \url{http://github.com/google/jax}.

\bibitem[Brandstetter et~al.(2022)Brandstetter, Worrall, and Welling]{brandstetter2022message}
Johannes Brandstetter, Daniel Worrall, and Max Welling.
\newblock Message passing neural pde solvers.
\newblock \emph{arXiv preprint arXiv:2202.03376}, 2022.

\bibitem[Brunton \& Kutz(2022)Brunton and Kutz]{brunton2022data}
Steven~L Brunton and J~Nathan Kutz.
\newblock \emph{Data-driven science and engineering: Machine learning, dynamical systems, and control}.
\newblock Cambridge University Press, 2022.

\bibitem[Bryutkin et~al.(2024)Bryutkin, Huang, Deng, Yang, Sch{\"o}nlieb, and Aviles-Rivero]{bryutkin2024hamlet}
Andrey Bryutkin, Jiahao Huang, Zhongying Deng, Guang Yang, Carola-Bibiane Sch{\"o}nlieb, and Angelica Aviles-Rivero.
\newblock Hamlet: Graph transformer neural operator for partial differential equations.
\newblock \emph{arXiv preprint arXiv:2402.03541}, 2024.

\bibitem[Cai et~al.(2021)Cai, Mao, Wang, Yin, and Karniadakis]{cai2021physics}
Shengze Cai, Zhiping Mao, Zhicheng Wang, Minglang Yin, and George~Em Karniadakis.
\newblock Physics-informed neural networks (pinns) for fluid mechanics: A review.
\newblock \emph{Acta Mechanica Sinica}, 37\penalty0 (12):\penalty0 1727--1738, 2021.

\bibitem[Capizzano(2003)]{capizzano2003generalized}
S~Serra Capizzano.
\newblock Generalized locally toeplitz sequences: spectral analysis and applications to discretized partial differential equations.
\newblock \emph{Linear Algebra and its Applications}, 366:\penalty0 371--402, 2003.

\bibitem[Carr \& Turner(2016)Carr and Turner]{carr2016semi}
EJ~Carr and IW~Turner.
\newblock A semi-analytical solution for multilayer diffusion in a composite medium consisting of a large number of layers.
\newblock \emph{Applied Mathematical Modelling}, 40\penalty0 (15-16):\penalty0 7034--7050, 2016.

\bibitem[Christie \& Blunt(2001)Christie and Blunt]{christie2001tenth}
Michael~Andrew Christie and Martin~J Blunt.
\newblock Tenth spe comparative solution project: A comparison of upscaling techniques.
\newblock \emph{SPE Reservoir Evaluation \& Engineering}, 4\penalty0 (04):\penalty0 308--317, 2001.

\bibitem[DeepMind et~al.(2020)DeepMind, Babuschkin, Baumli, Bell, Bhupatiraju, Bruce, Buchlovsky, Budden, Cai, Clark, Danihelka, Dedieu, Fantacci, Godwin, Jones, Hemsley, Hennigan, Hessel, Hou, Kapturowski, Keck, Kemaev, King, Kunesch, Martens, Merzic, Mikulik, Norman, Papamakarios, Quan, Ring, Ruiz, Sanchez, Sartran, Schneider, Sezener, Spencer, Srinivasan, Stanojevi\'{c}, Stokowiec, Wang, Zhou, and Viola]{deepmind2020jax}
DeepMind, Igor Babuschkin, Kate Baumli, Alison Bell, Surya Bhupatiraju, Jake Bruce, Peter Buchlovsky, David Budden, Trevor Cai, Aidan Clark, Ivo Danihelka, Antoine Dedieu, Claudio Fantacci, Jonathan Godwin, Chris Jones, Ross Hemsley, Tom Hennigan, Matteo Hessel, Shaobo Hou, Steven Kapturowski, Thomas Keck, Iurii Kemaev, Michael King, Markus Kunesch, Lena Martens, Hamza Merzic, Vladimir Mikulik, Tamara Norman, George Papamakarios, John Quan, Roman Ring, Francisco Ruiz, Alvaro Sanchez, Laurent Sartran, Rosalia Schneider, Eren Sezener, Stephen Spencer, Srivatsan Srinivasan, Milo\v{s} Stanojevi\'{c}, Wojciech Stokowiec, Luyu Wang, Guangyao Zhou, and Fabio Viola.
\newblock The {D}eep{M}ind {JAX} {E}cosystem, 2020.
\newblock URL \url{http://github.com/google-deepmind}.

\bibitem[Dulny et~al.(2023)Dulny, Hotho, and Krause]{dulny2023dynabench}
Andrzej Dulny, Andreas Hotho, and Anna Krause.
\newblock Dynabench: A benchmark dataset for learning dynamical systems from low-resolution data.
\newblock In \emph{Joint European Conference on Machine Learning and Knowledge Discovery in Databases}, pp.\  438--455. Springer, 2023.

\bibitem[Eivazi et~al.(2024)Eivazi, Wang, and Vinuesa]{eivazi2024physics}
Hamidreza Eivazi, Yuning Wang, and Ricardo Vinuesa.
\newblock Physics-informed deep-learning applications to experimental fluid mechanics.
\newblock \emph{Measurement science and technology}, 35\penalty0 (7):\penalty0 075303, 2024.

\bibitem[Elman et~al.(2014)Elman, Silvester, and Wathen]{elman2014finite}
Howard~C Elman, David~J Silvester, and Andrew~J Wathen.
\newblock \emph{Finite elements and fast iterative solvers: with applications in incompressible fluid dynamics}.
\newblock Oxford university press, 2014.

\bibitem[Eymard et~al.(2000)Eymard, Gallou{\"e}t, and Herbin]{eymard2000finite}
Robert Eymard, Thierry Gallou{\"e}t, and Rapha{\`e}le Herbin.
\newblock Finite volume methods.
\newblock \emph{Handbook of numerical analysis}, 7:\penalty0 713--1018, 2000.

\bibitem[Fanaskov \& Oseledets(2023)Fanaskov and Oseledets]{fanaskov2023spectral}
VS~Fanaskov and Ivan~V Oseledets.
\newblock Spectral neural operators.
\newblock In \emph{Doklady Mathematics}, volume 108, pp.\  S226--S232. Springer, 2023.

\bibitem[Gupta \& Brandstetter(2022)Gupta and Brandstetter]{gupta2022towards}
Jayesh~K Gupta and Johannes Brandstetter.
\newblock Towards multi-spatiotemporal-scale generalized pde modeling.
\newblock \emph{arXiv preprint arXiv:2209.15616}, 2022.

\bibitem[Hao et~al.(2023)Hao, Yao, Su, Su, Wang, Lu, Xia, Zhang, Liu, Lu, et~al.]{hao2023pinnacle}
Zhongkai Hao, Jiachen Yao, Chang Su, Hang Su, Ziao Wang, Fanzhi Lu, Zeyu Xia, Yichi Zhang, Songming Liu, Lu~Lu, et~al.
\newblock Pinnacle: A comprehensive benchmark of physics-informed neural networks for solving pdes.
\newblock \emph{arXiv preprint arXiv:2306.08827}, 2023.

\bibitem[Hashmi(2014)]{hashmi2014comprehensive}
M~Saleem~J Hashmi.
\newblock \emph{Comprehensive materials processing}.
\newblock Newnes, 2014.

\bibitem[Hassan et~al.(2023)Hassan, Feeney, Dhruv, Kim, Suh, Ryu, Won, and Chandramowlishwaran]{hassan2023bubbleml}
Sheikh Md~Shakeel Hassan, Arthur Feeney, Akash Dhruv, Jihoon Kim, Youngjoon Suh, Jaiyoung Ryu, Yoonjin Won, and Aparna Chandramowlishwaran.
\newblock Bubbleml: A multi-physics dataset and benchmarks for machine learning.
\newblock \emph{arXiv preprint arXiv:2307.14623}, 2023.

\bibitem[Holl et~al.(2020)Holl, Koltun, and Thuerey]{holl2020learning}
Philipp Holl, Vladlen Koltun, and Nils Thuerey.
\newblock Learning to control pdes with differentiable physics.
\newblock \emph{arXiv preprint arXiv:2001.07457}, 2020.

\bibitem[Hsieh et~al.(2019)Hsieh, Zhao, Eismann, Mirabella, and Ermon]{hsieh2019learning}
Jun-Ting Hsieh, Shengjia Zhao, Stephan Eismann, Lucia Mirabella, and Stefano Ermon.
\newblock Learning neural pde solvers with convergence guarantees.
\newblock \emph{arXiv preprint arXiv:1906.01200}, 2019.

\bibitem[Hu et~al.(2019)Hu, Anderson, Li, Sun, Carr, Ragan-Kelley, and Durand]{hu2019difftaichi}
Yuanming Hu, Luke Anderson, Tzu-Mao Li, Qi~Sun, Nathan Carr, Jonathan Ragan-Kelley, and Fr{\'e}do Durand.
\newblock Difftaichi: Differentiable programming for physical simulation.
\newblock \emph{arXiv preprint arXiv:1910.00935}, 2019.

\bibitem[Ingraham et~al.(2018)Ingraham, Riesselman, Sander, and Marks]{ingraham2018learning}
John Ingraham, Adam Riesselman, Chris Sander, and Debora Marks.
\newblock Learning protein structure with a differentiable simulator.
\newblock In \emph{International conference on learning representations}, 2018.

\bibitem[Johnson et~al.(2024)Johnson, Febvre, Gorbunova, Metref, Ballarotta, Le~Sommer, et~al.]{johnson2024oceanbench}
J~Emmanuel Johnson, Quentin Febvre, Anastasiia Gorbunova, Sam Metref, Maxime Ballarotta, Julien Le~Sommer, et~al.
\newblock Oceanbench: The sea surface height edition.
\newblock \emph{Advances in Neural Information Processing Systems}, 36, 2024.

\bibitem[Karniadakis et~al.(2021)Karniadakis, Kevrekidis, Lu, Perdikaris, Wang, and Yang]{karniadakis2021physics}
George~Em Karniadakis, Ioannis~G Kevrekidis, Lu~Lu, Paris Perdikaris, Sifan Wang, and Liu Yang.
\newblock Physics-informed machine learning.
\newblock \emph{Nature Reviews Physics}, 3\penalty0 (6):\penalty0 422--440, 2021.

\bibitem[Kidger \& Garcia(2021)Kidger and Garcia]{kidger2021equinox}
Patrick Kidger and Cristian Garcia.
\newblock {E}quinox: neural networks in {JAX} via callable {P}y{T}rees and filtered transformations.
\newblock \emph{Differentiable Programming workshop at Neural Information Processing Systems 2021}, 2021.

\bibitem[LeVeque(2007)]{leveque2007finite}
Randall~J LeVeque.
\newblock \emph{Finite difference methods for ordinary and partial differential equations: steady-state and time-dependent problems}.
\newblock SIAM, 2007.

\bibitem[Li et~al.(2020)Li, Kovachki, Azizzadenesheli, Liu, Bhattacharya, Stuart, and Anandkumar]{li2020fourier}
Zongyi Li, Nikola Kovachki, Kamyar Azizzadenesheli, Burigede Liu, Kaushik Bhattacharya, Andrew Stuart, and Anima Anandkumar.
\newblock Fourier neural operator for parametric partial differential equations.
\newblock \emph{arXiv preprint arXiv:2010.08895}, 2020.

\bibitem[Li et~al.(2024)Li, Kovachki, Choy, Li, Kossaifi, Otta, Nabian, Stadler, Hundt, Azizzadenesheli, et~al.]{li2024geometry}
Zongyi Li, Nikola Kovachki, Chris Choy, Boyi Li, Jean Kossaifi, Shourya Otta, Mohammad~Amin Nabian, Maximilian Stadler, Christian Hundt, Kamyar Azizzadenesheli, et~al.
\newblock Geometry-informed neural operator for large-scale 3d pdes.
\newblock \emph{Advances in Neural Information Processing Systems}, 36, 2024.

\bibitem[Lu et~al.(2021{\natexlab{a}})Lu, Jin, Pang, Zhang, and Karniadakis]{lu2021learning}
Lu~Lu, Pengzhan Jin, Guofei Pang, Zhongqiang Zhang, and George~Em Karniadakis.
\newblock Learning nonlinear operators via deeponet based on the universal approximation theorem of operators.
\newblock \emph{Nature machine intelligence}, 3\penalty0 (3):\penalty0 218--229, 2021{\natexlab{a}}.

\bibitem[Lu et~al.(2021{\natexlab{b}})Lu, Meng, Mao, and Karniadakis]{lu2021deepxde}
Lu~Lu, Xuhui Meng, Zhiping Mao, and George~Em Karniadakis.
\newblock {DeepXDE}: A deep learning library for solving differential equations.
\newblock \emph{SIAM Review}, 63\penalty0 (1):\penalty0 208--228, 2021{\natexlab{b}}.
\newblock \doi{10.1137/19M1274067}.

\bibitem[Luo et~al.(2023)Luo, Chen, and Zhang]{luo2023cfdbench}
Yining Luo, Yingfa Chen, and Zhen Zhang.
\newblock Cfdbench: A comprehensive benchmark for machine learning methods in fluid dynamics.
\newblock \emph{arXiv preprint arXiv:2310.05963}, 2023.

\bibitem[Ma et~al.(2021)Ma, Zhang, Thuerey, Hu, and Haidn]{ma2021physics}
Hao Ma, Yuxuan Zhang, Nils Thuerey, Xiangyu Hu, and Oskar~J Haidn.
\newblock Physics-driven learning of the steady navier-stokes equations using deep convolutional neural networks.
\newblock \emph{arXiv preprint arXiv:2106.09301}, 2021.

\bibitem[Massimo(2013)]{massimo2013physics}
Luigi Massimo.
\newblock \emph{Physics of high-temperature reactors}.
\newblock Elsevier, 2013.

\bibitem[Muravleva et~al.(2021)Muravleva, Derbyshev, Boronin, and Osiptsov]{muravleva2021multigrid}
Ekaterina~A Muravleva, Dmitry~Yu Derbyshev, Sergei~A Boronin, and Andrei~A Osiptsov.
\newblock Multigrid pressure solver for 2d displacement problems in drilling, cementing, fracturing and eor.
\newblock \emph{Journal of Petroleum Science and Engineering}, 196:\penalty0 107918, 2021.

\bibitem[Nathaniel et~al.(2024)Nathaniel, Qu, Nguyen, Yu, Busecke, Grover, and Gentine]{nathaniel2024chaosbench}
Juan Nathaniel, Yongquan Qu, Tung Nguyen, Sungduk Yu, Julius Busecke, Aditya Grover, and Pierre Gentine.
\newblock Chaosbench: A multi-channel, physics-based benchmark for subseasonal-to-seasonal climate prediction.
\newblock \emph{arXiv preprint arXiv:2402.00712}, 2024.

\bibitem[Nguyen et~al.(2023)Nguyen, Vu, Nguyen, Huynh, Nguyen, and Hy]{nguyen2023neural}
Duc~Minh Nguyen, Minh~Chau Vu, Tuan~Anh Nguyen, Tri Huynh, Nguyen~Tri Nguyen, and Truong~Son Hy.
\newblock Neural multigrid memory for computational fluid dynamics.
\newblock \emph{arXiv preprint arXiv:2306.12545}, 2023.

\bibitem[Oristaglio \& Hohmann(1984)Oristaglio and Hohmann]{oristaglio1984diffusion}
Michael~L Oristaglio and Gerald~W Hohmann.
\newblock Diffusion of electromagnetic fields into a two-dimensional earth: A finite-difference approach.
\newblock \emph{Geophysics}, 49\penalty0 (7):\penalty0 870--894, 1984.

\bibitem[Raissi et~al.(2019)Raissi, Perdikaris, and Karniadakis]{raissi2019physics}
Maziar Raissi, Paris Perdikaris, and George~E Karniadakis.
\newblock Physics-informed neural networks: A deep learning framework for solving forward and inverse problems involving nonlinear partial differential equations.
\newblock \emph{Journal of Computational physics}, 378:\penalty0 686--707, 2019.

\bibitem[Ren et~al.(2023)Ren, Erichson, Subramanian, San, Lukic, and Mahoney]{ren2023superbench}
Pu~Ren, N~Benjamin Erichson, Shashank Subramanian, Omer San, Zarija Lukic, and Michael~W Mahoney.
\newblock Superbench: A super-resolution benchmark dataset for scientific machine learning.
\newblock \emph{arXiv preprint arXiv:2306.14070}, 2023.

\bibitem[Ripken et~al.(2023)Ripken, Coiffard, Pieper, and Dziadzio]{ripken2023multiscale}
Winfried Ripken, Lisa Coiffard, Felix Pieper, and Sebastian Dziadzio.
\newblock Multiscale neural operators for solving time-independent pdes.
\newblock \emph{arXiv preprint arXiv:2311.05964}, 2023.

\bibitem[Ronneberger et~al.(2015)Ronneberger, Fischer, and Brox]{ronneberger2015u}
Olaf Ronneberger, Philipp Fischer, and Thomas Brox.
\newblock U-net: Convolutional networks for biomedical image segmentation.
\newblock In \emph{Medical image computing and computer-assisted intervention--MICCAI 2015: 18th international conference, Munich, Germany, October 5-9, 2015, proceedings, part III 18}, pp.\  234--241. Springer, 2015.

\bibitem[Saad(2003)]{saad2003iterative}
Yousef Saad.
\newblock \emph{Iterative methods for sparse linear systems}.
\newblock SIAM, 2003.

\bibitem[Schnell \& Thuerey(2024)Schnell and Thuerey]{schnell2024stabilizing}
Patrick Schnell and Nils Thuerey.
\newblock Stabilizing backpropagation through time to learn complex physics.
\newblock \emph{arXiv preprint arXiv:2405.02041}, 2024.

\bibitem[Stachenfeld et~al.(2021)Stachenfeld, Fielding, Kochkov, Cranmer, Pfaff, Godwin, Cui, Ho, Battaglia, and Sanchez-Gonzalez]{stachenfeld2021learned}
Kimberly Stachenfeld, Drummond~B Fielding, Dmitrii Kochkov, Miles Cranmer, Tobias Pfaff, Jonathan Godwin, Can Cui, Shirley Ho, Peter Battaglia, and Alvaro Sanchez-Gonzalez.
\newblock Learned coarse models for efficient turbulence simulation.
\newblock \emph{arXiv preprint arXiv:2112.15275}, 2021.

\bibitem[Takamoto et~al.(2022)Takamoto, Praditia, Leiteritz, MacKinlay, Alesiani, Pfl{\"u}ger, and Niepert]{takamoto2022pdebench}
Makoto Takamoto, Timothy Praditia, Raphael Leiteritz, Daniel MacKinlay, Francesco Alesiani, Dirk Pfl{\"u}ger, and Mathias Niepert.
\newblock Pdebench: An extensive benchmark for scientific machine learning.
\newblock \emph{Advances in Neural Information Processing Systems}, 35:\penalty0 1596--1611, 2022.

\bibitem[Tali et~al.(2024)Tali, Rabeh, Yang, Shadkhah, Karki, Upadhyaya, Dhakshinamoorthy, Saadati, Sarkar, Krishnamurthy, et~al.]{tali2024flowbench}
Ronak Tali, Ali Rabeh, Cheng-Hau Yang, Mehdi Shadkhah, Samundra Karki, Abhisek Upadhyaya, Suriya Dhakshinamoorthy, Marjan Saadati, Soumik Sarkar, Adarsh Krishnamurthy, et~al.
\newblock Flowbench: A large scale benchmark for flow simulation over complex geometries.
\newblock \emph{arXiv preprint arXiv:2409.18032}, 2024.

\bibitem[Toshev et~al.(2024)Toshev, Galletti, Fritz, Adami, and Adams]{toshev2024lagrangebench}
Artur Toshev, Gianluca Galletti, Fabian Fritz, Stefan Adami, and Nikolaus Adams.
\newblock Lagrangebench: A lagrangian fluid mechanics benchmarking suite.
\newblock \emph{Advances in Neural Information Processing Systems}, 36, 2024.

\bibitem[Tran et~al.(2021)Tran, Mathews, Xie, and Ong]{tran2021factorized}
Alasdair Tran, Alexander Mathews, Lexing Xie, and Cheng~Soon Ong.
\newblock Factorized fourier neural operators.
\newblock \emph{arXiv preprint arXiv:2111.13802}, 2021.

\bibitem[Trefethen(2000)]{trefethen2000spectral}
Lloyd~N Trefethen.
\newblock \emph{Spectral methods in MATLAB}.
\newblock SIAM, 2000.

\bibitem[Wang et~al.(2020)Wang, Walters, and Yu]{wang2020incorporating}
Rui Wang, Robin Walters, and Rose Yu.
\newblock Incorporating symmetry into deep dynamics models for improved generalization.
\newblock \emph{arXiv preprint arXiv:2002.03061}, 2020.

\bibitem[Yu et~al.(2017)Yu, Koltun, and Funkhouser]{yu2017dilated}
Fisher Yu, Vladlen Koltun, and Thomas Funkhouser.
\newblock Dilated residual networks.
\newblock In \emph{Proceedings of the IEEE conference on computer vision and pattern recognition}, pp.\  472--480, 2017.

\bibitem[Yu et~al.(2024)Yu, Hannah, Peng, Lin, Bhouri, Gupta, L{\"u}tjens, Will, Behrens, Busecke, et~al.]{yu2024climsim}
Sungduk Yu, Walter Hannah, Liran Peng, Jerry Lin, Mohamed~Aziz Bhouri, Ritwik Gupta, Bj{\"o}rn L{\"u}tjens, Justus~C Will, Gunnar Behrens, Julius Busecke, et~al.
\newblock Climsim: A large multi-scale dataset for hybrid physics-ml climate emulation.
\newblock \emph{Advances in Neural Information Processing Systems}, 36, 2024.

\end{thebibliography}
\bibliographystyle{iclr2025_conference}

\newpage
\appendix
\section{Appendix}

\subsection{Architectures}
\label{app:models}

In this section, we discuss the architectures used in more detail and provide information on the training procedures and hyperparameters used. The list of used models is:

\begin{enumerate}
    \item F-FNO -- Factorized Fourier Neural Operator (F-FNO) from \citep{tran2021factorized}.
    \item fSNO -- Spectral Neural Operator (SNO). The construction mirrors FNO, but instead of FFT, a transformation based on Gauss quadratures is used \citep{fanaskov2023spectral}.
    \item DilResNet -- Dilated Residual Network from \citep{yu2017dilated}, \citep{stachenfeld2021learned}.
    \item U-Net -- classical computer vision architecture introduced in \citep{ronneberger2015u}.
\end{enumerate}

\paragraph{F-FNO}

Unlike the original \citep{li2020fourier}, the authors of \citep{tran2021factorized} proposed to changing the operator layer to:

\begin{equation*}
    z^{\ell+1} = z^{\ell} + \sigma \Big[W_2^{(\ell)}\sigma\Big(W_1^{(\ell)}\mathcal{K}^{(\ell)}(z^{(\ell)}) + b_1^{(\ell)}\Big) + b_2^{(\ell)}\Big],
\end{equation*}

\noindent where $\sigma$ is an activation function, $W_1$ and $W_2$ are weight matrices  in the physical space, $b_1$ and $b_2$ are bias vectors and

\begin{equation*}
    \mathcal{K}^{(\ell)}\big(z^{(\ell)}\big) = \sum_{d \in D}\Big[\text{IFFT}\big(R_d^{(\ell)} \cdot \text{FFT}_d(z^{\ell})\big)\Big],
\end{equation*}

\noindent where $R_d$ is a Fourier domain weight matrix, FFT and IFFT are Fast Fourier and inverse Fast Fourier transforms.

F-FNO has an encoder-processor-decoder architecture. We used the following parameters: $4$ Fourier layers in the processor, $12$ modes and $\text{GeLU}$ as the activation function. We used $48$ features in the processor.


\paragraph{SNO}

We utilized spectral neural operators (SNO) \citep{fanaskov2023spectral} with linear integral kernels:

\begin{equation*}
    u\leftarrow \int dx A_{ij} p_j(x) \left(p_i, u\right)~,
\end{equation*}

\noindent where $p_j(x)$ are orthogonal or trigonometric polynomials. 

These linear integral kernels are an extension of the integral kernels used in the FNO \citep{li2020fourier}. More specifically, starting from the input function $u^{n}$, we produce the output function $u^{n+1}$, which is later transformed by nonlinear activation. The transformation depends on the set of polynomials $p_{j}$ that form a suitable basis for the problem at hand (e.g. trigonometric polynomials, Chebyshev polynomials, etc.). These polynomials are chosen beforehand and do not change during training. The transformation is naturally divided into three parts: analysis, processing, synthesis.

At the analysis stage, we find a discrete representation of the input function by projecting it onto a set of polynomials. To do this, we compute scalar products:

\begin{equation*}
    \alpha_{j} = \left(p_{j}, u^{n}\right) = \int dx p_{j}(x) u^{n}(x) w(x)~,
\end{equation*}

\noindent where $w(x)$ is a non-negative weight function given by the polynomial used.

At the processing stage, we process the obtained coefficients with a linear layer:

\begin{equation*}
    \alpha_{i}^{‘} = \sum_{j}A_{ij} \alpha_{j}~.
\end{equation*}

Finally, at the synthesis stage, we recover the continuous function as the sum of the processed coefficients:

\begin{equation*}
    u^{n+1} = \sum_{j}p_{j} \alpha_{j}^{‘}.
\end{equation*}

We use SNO in Fourier basis (see \citep{fanaskov2023spectral}) with encoder-processor-decoder architecture. The number of SNO layers is $4$ and the number of $p_j(x)$ is $20$. We use $\text{GeLU}$ as activation function. 

\paragraph{DilResNet}

The conventional dilated residual network was first proposed in \citep{stachenfeld2021learned}. In this study, the DilResNet architecture is configured with four blocks, each consisting of a sequence of convolutions with steps of $[1, 2, 4, 8, 4, 2, 1]$ and a kernel size of $3$. Skip connections are also applied after each block and the $\text{GeLU}$ activation function is used.

\paragraph{U-Net}

We adopt the traditional U-Net architecture proposed in \citep{ronneberger2015u}. This U-Net configuration is characterised by a series of levels, where each level has approximately half the resolution of the previous one, and the number of features is doubled. At each level, we apply a sequence of three convolutions, followed by max pooling, and then a transposed convolution for upsampling. After upsampling, three more convolutions are applied at each level. The U-Net used in this study consists of four layers and incorporates the $\text{GeLU}$ activation function.

\end{document}